# A Generalized Meta-loss function for regression and classification using privileged information

Amina Asif, Muhammad Dawood, Fayyaz ul Amir Afsar Minhas*

*PIEAS Data Science Lab*
*Pakistan Institute of Engineering and Applied Sciences (PIEAS),*
*PO Nilore, Islamabad, Pakistan.*
*\* corresponding author email: afsar@pieas.edu.pk*

**Abstract**

Learning using privileged information (LUPI) is a powerful heterogenous feature space machine learning framework that allows a machine learning model to learn from highly informative or *privileged* features which are available during training only to generate test predictions using *input* space features which are available both during training and testing. LUPI can significantly improve prediction performance in a variety of machine learning problems. However, existing large margin and neural network implementations of learning using privileged information are mostly designed for classification tasks. In this work, we have proposed a simple yet effective formulation that allows us to perform regression using privileged information through a custom loss function. Apart from regression, our formulation allows general application of LUPI to classification and other related problems as well. We have verified the correctness, applicability and effectiveness of our method on regression and classification problems over different synthetic and real-world problems. To test the usefulness of the proposed model in real-world problems, we have evaluated our method on the problem of protein binding affinity prediction. The proposed LUPI regression-based model has shown to outperform the current state-of-the-art predictor.

*Keywords[1]:* Learning Using Privileged Information; Distillation; Loss function; Regression; Classification; Neural Networks

## 1. Introduction

Machine learning has seen some major advancements over the past few decades [1], [2]. Supervised learning is one of the major areas of interest for scientists in the field. In conventional supervised learning, a model is first trained over some training examples and then deployed for real-world use after testing [3]. Every model is trained using some features which are either hand-crafted or, in the case of many deep learning applications, extracted automatically from given examples. During testing, the same features are extracted from the test examples for generating predictions over these examples. That is, in regular supervised learning, training and testing are performed in the same feature space. One of the limitations of such methods is that even if we have access to some very important and informative features for training examples, they cannot be used for training the model unless the same features can be extracted for test examples as well. Learning methods over different or heterogeneous spaces in training and testing are needed in many practical scenarios. Consider the development of an automated medical diagnosis system using features derived from a patient's current condition (such as temperature, histopathology images, etc.) as input to diagnose his/her disease. For training, the developer may have access to other important knowledge as well that may only be available for training examples, e.g., the doctor's notes, recovery information, condition after a surgery, etc. However, such information can be invaluable for developing such a predictor. Another such example from the domain of bioinformatics is the development of predictive models for certain properties of proteins such as binding affinity, interaction site, etc. Amino acid sequence information is more readily available for proteins in comparison to their 3-D structures [4]. Therefore, a predictor that uses only sequence information for prediction may be of great value to biologists. However, structural features of proteins are more informative and may be available for several training examples. Conventional supervised machine learning methods cannot use the more informative features if they are available during training only. A predictor that can generate predictions using sequence information alone but utilize the information contain in protein structures during its training can be expected to perform better than a predictor that does not use structure information.

---
[1] An original copy of the paper is available at arXiv: https://arxiv.org/abs/1811.06885

To overcome the above-mentioned limitations, Vapnik et al. proposed a new learning paradigm called Learning Using Privileged Information (LUPI) [5], [6] for scenarios where certain features, called *privileged information*, are available during training only, whereas, *input space* features are available both for training and testing examples. Using LUPI, privileged information can be used during training along with regular input space features to get a better classifier in the input space that can generate predictions using input space features alone. Results have shown that LUPI based classification models can perform better than models trained using input space features only in a number of machine learning problems including image and object classification [7]–[11].

LUPI has been formulated for a number of learning methods such as Support Vector Machines (SVMs) [6], [12], [13], Neural Networks [14], [15], Extreme Learning Machines [16], Random Forests[17], etc. However, existing formulations of LUPI based large-margin models and neural networks in the literature have been shown to work for classification tasks only. As discussed in more detail in the next section, this limitation is primarily due to the use of label softening as a means of regularization to control the flow of information between privileged and input feature spaces [18]. In this work, we propose a very simple but generalized formulation that can be used to perform regression using privileged information. Apart from regression, our formulation can also be applied effectively for other machine learning tasks such as classification and ranking as well. We provide a complete mathematical formulation as well as implementation of the proposed scheme. We have also performed extensive performance assessment of the proposed scheme over both toy datasets as well as a practical problem of predicting protein binding affinity from sequence and structure information. It is important to note that we have presented performance results for both classification and regression tasks and to the best of the knowledge of these authors this is a first generalized implementation of learning using privileged information that can be integrated with neural network models for practically any type of machine learning problem.

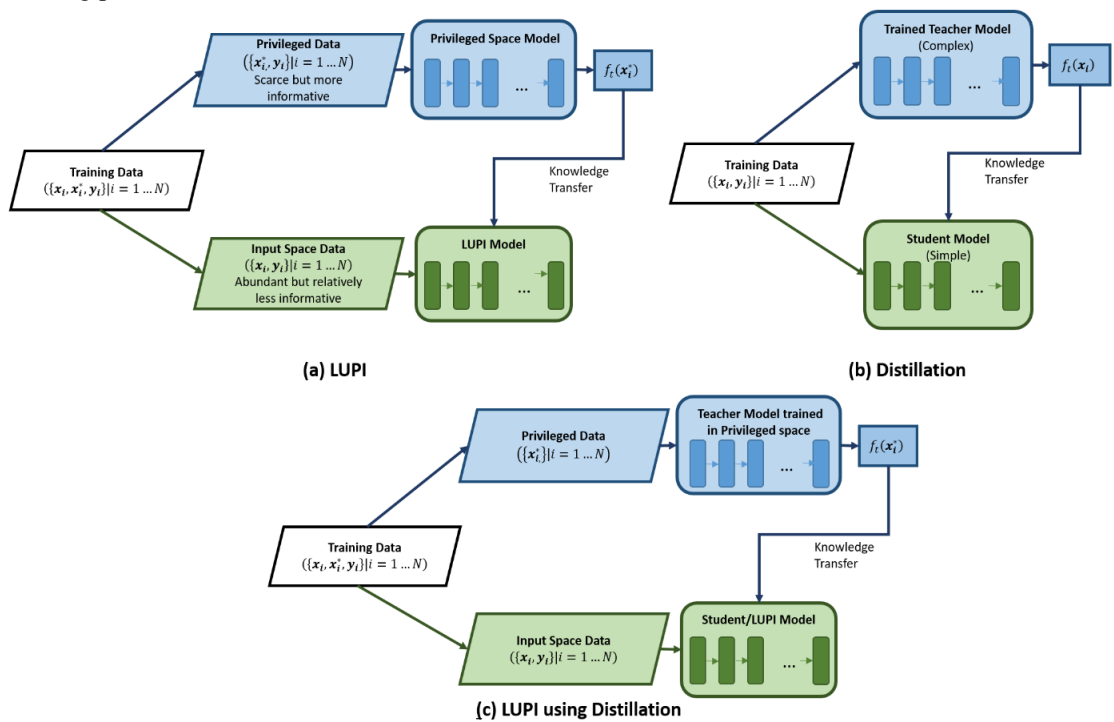

Figure 1 Illustration of concepts of LUPI, distillation and LUPI using distillation

## 2. Methods

In this section, we present the mathematical formulations and details of different experiments carried out to evaluate our method.

*2.1. LUPI and Distillation*

Our formulation exploits the recent work by Lopez-Paz et al. which unifies LUPI based classification with another machine learning problem called distillation [19]. Before delving into the details of the proposed scheme, we begin by a description of the equivalence of LUPI and distillation. Distillation is inspired from problems associated with the deployment of computationally intensive DNN models in resource-limited environments like mobile devices [20],[18]. In distillation, a computationally intensive predictive model called the teacher is used for training a smaller and simpler model called the student [18] by reducing the loss between the predictions generated by the two models, i.e., making a student learn from the teacher model. Distillation has been shown to be effective in a variety of problems in computer vision, image processing and other domains [21]–[25]. Lopez-Paz et al. have provided a proof of unification of the concepts of LUPI and distillation: a teacher model trained over privileged information can be used to transfer knowledge to a student model in the input feature space. Thus, distillation from *privileged* space to *input* space is equivalent to LUPI (see Figure 1). In order to formulate the problem of LUPI mathematically and understand the limitations associated with it, consider a dataset of $N$ training examples given by $\{(x_i, x_i^*, y_i)|i = 1 \dots N\}$ with input space features ($x_i$), privileged space feature ($x_i^*$) and target values $y_i$ for examples $i = 1 \dots N$. LUPI works by first training a teacher model $f_t(x_i^*) \in \mathcal{F}_t$ in the privileged space and then distilling an input space or student model $f_s(x_i) \in \mathcal{F}_s$ from the teacher model. The student model can then be used for generating predictions for test examples in the input feature space. Knowledge transfer from the teacher to the student is achieved by utilizing a meta-loss function that penalizes errors of the student model in comparison with the actual target value as well as the prediction of the teacher for a given example. Given a non-negative loss function $l(y, u)$ that gives the error of the prediction $u$ of a model and the target $y$, the learning problem of distillation based LUPI [19] (see Fig. 1) can be expressed as the following minimization of a meta loss function involving two loss terms: one with the prediction of the teacher and the other with the target value:

$$f_s = \arg min_{f \in \mathcal{F}_s} \frac{1}{N} \sum_{i=1}^{N} \left[ (1-\lambda) l\left(y_i, \sigma(f(x_i))\right) + \lambda l\left(\sigma(f_t(x_i^*)/T), \sigma(f(x_i))\right) \right] \quad (1)$$

Here, $l\left(y_i, \sigma(f(x_i))\right)$ represents the loss between the actual target value $y_i$ and the student's output $f(x_i)$, $l\left(\sigma(f_t(x_i^*)/T), \sigma(f(x_i))\right)$ is the loss between *softened* predictions by the teacher model and the student's prediction. $\sigma(\cdot)$ is the soft-max function and the hyper-parameter $\lambda$ ($0 \leq \lambda \leq 1$) controls the trade-off between learning directly from the target values and learning from the teacher. The division of the teacher's prediction by a temperature hyper-parameter $T > 0$ in teacher-student loss is called softening and is performed to smooth out the predicted class probabilities. This, in effect, acts as a regularization factor and can be used to control transfer of knowledge from the teacher to the student. The objective of softening can be interpreted as preventing the student from mimicking the teacher if the teacher is in error. However, the use of division for softening in the above formulation prevents the application of this approach directly in regression problems as dividing by $T$ and applying soft-max becomes meaningless in problems other than classification. This limits the applicability of LUPI to problems such as regression, ranking or the design of recommender systems [19], [24].

*2.2. Proposed Formulation*

As discussed above, the objective of LUPI and distillation is to transfer *useful* knowledge from the teacher to the student. Thus, a student model should learn from the teacher if the teacher's predictions are reliable and otherwise the student should learn from the given labels directly. Equation (1) achieves this by softening the outputs of the teacher model by application of the soft-max function and division by a temperature factor $T$ which limits its use to classification problems only as such operations are not meaningful in other types of machine learning problems such as regression. We now present a novel and simple but very effective modification to equation (1) to resolve the issues highlighted in the above section. Our proposed scheme achieves the desired control of knowledge transfer from the teacher to the student by weighing the loss between the outputs of the student and the teacher $l\left(f(x_i), f_t(x_i^*)\right)$ by a factor $e^{-Tl(f_t(x_i^*),y_i)}$ which is inversely related to the loss between the teacher's output and the actual targets $l(f_t(x_i^*), y_i)$. The complete mathematical formulation becomes:

$$f_s = \arg min_{f \in \mathcal{F}_s} \frac{1}{N} \sum_{i=1}^{N} \left(1 - e^{-Tl(f_t(x_i^*), y_i)}\right) l(f(x_i), y_i) + e^{-Tl(f_t(x_i^*), y_i)} l(f(x_i), f_t(x_i^*)) \quad (2)$$

Here, $T$ is the temperature hyper-parameter that controls the extent to which student should mimic the teacher. For small values of $T$ such as $T = 0$ or when the teacher's prediction are close to the target value with small $l(f_t(x_i^*), y_i)$, the student will try to mimic the teacher by putting more emphasis on the reduction of $l(f(x_i), f_t(x_i^*))$ in comparison to $l(f(x_i), y_i)$ and, as a result, student predictions will be close to the teacher's output. For large values of $T$ such as $T \rightarrow \infty$ or when the loss $l(f_t(x_i^*), y_i)$ is large, there would be limited or no knowledge transfer from the teacher and the student model will try to learn by minimizing the empirical loss between its outputs and the target value only. Consequently, the input space model will learn from the privileged space model only when predictions from the teacher (privileged space) are more reliable. Thus, the proposed model can achieve the same effect as softening for knowledge transfer as equation (1) but without involving any soft-max or division of the model's output by $T$ as in equation (1). This makes the proposed formulation more generalized and amenable to several machine learning problems like classification, ranking, regression, etc. The degree of knowledge transfer can be controlled by a single hyper-parameter $T$ which can be set through cross-validation. Furthermore, it is also possible to add domain knowledge by setting $T$ for individual examples to reflect our confidence in the given targets or the correctness of predictions generated by the teacher model.

*2.3. Experiments*

To assess the correctness and effectiveness of our method, we have evaluated its performance on three different types of datasets. All the teacher/student models have been implemented as multi-layered perceptron neural networks in pyTorch [26]. The code for reconstructing all experiments is available at the URL: https://gitlab.com/muhammad.dawood/lupi_pytorch.

The three different dataset categories over which evaluated our method are: 1) the synthetic data used by Lopez-Paz et al. [19], 2) MNIST number classification, and 3) protein binding affinity dataset. Lopez-Paz et al. [19] have performed experiments for classification only. Our focus in this study is to solve regression problems using privileged information, but to prove the applicability of the proposed formulation in different machine learning tasks, we have performed analysis in both classification and regression settings for the synthetic datasets. To further prove its effectiveness in real-world classification problems, we have evaluated the proposed method over the MNIST classification dataset. We have also tested our method over a real-world LUPI problem of predicting protein binding affinity. Details of the experiments are presented in the following subsections.

*2.3.1. Synthetic Datasets*

As discussed earlier, to prove the correctness of our formulation we have tested the method over artificially generated datasets by Lopez-Paz et al. [19]. The study in [19] has performed experiments for classification only. Since our goal is to solve regression problems as well using LUPI, we assign continuous labels to the artificially created data samples and evaluate our method over the resulting regression problem. Also, to prove that our method is equally effective in classification problems, we have applied and analyzed its performance on the synthetic datasets in classification settings as well. Four experiments using the artificial data have been carried out, each representing a possible privileged-input space configuration. In each of the experiments, 50-dimensional input space feature vectors drawn from normal distribution are used. The examples are projected onto a hyperplane sampled from a normal distribution to generate target values. For each of these experiments, 200 training examples and 1000 test examples are used. To perform a fair performance comparison with [19], a simple single layer neural network with linear activations is used for each dataset. We have used Binary Cross-Entropy (BCE) and Mean Squared Error (MSE) losses for classification and regression, respectively. Each experiment is repeated 10 times and the average and standard deviation of accuracy (for classification) and Root Mean Squared Error (RMSE) (for regression) is reported [27]. Further details of each of the privileged-input space configuration employed is given as follows.

*2.3.1.1. Clean Labels as Privileged Information*

In this experiment, noise-free true targets are used as privileged information while some noise is added to the targets. Mathematically,

$$x_i \sim \aleph(0, I_d), d = 50$$
$$\alpha \sim \aleph(0, I_d)$$
$$x_i^* \leftarrow \alpha^T x_i$$
$$\varepsilon_i \sim \aleph(0, 1)$$

For classification, the true labels are obtained using the indicator function $\mathbb{I}(\cdot)$:
$$y_i \leftarrow \mathbb{I}((x_i^* + \varepsilon_i) > 0)$$

For regression, the target values are set as:
$$y_i \leftarrow (x_i^* + \varepsilon_i)$$

*2.3.1.2. Clean Features as Privileged Information*

Here, we use noisy features as input space samples and pass clean features as privileged space features. Mathematically,
$$x_i^* \sim \aleph(0, I_d), d = 50$$
$$\alpha \sim \aleph(0, I_d)$$
$$\varepsilon_i \sim \aleph(0, I_d)$$
$$x_i \leftarrow (x_i^* + \varepsilon_i)$$

For classification, the targets are obtained as:
$$y_i \leftarrow \mathbb{I}(\alpha^T x_i > 0)$$

For regression, the targets are given by:
$$y_i \leftarrow \alpha^T x_i$$

*2.3.1.3. Relevant Features as Privileged Information*

Here, we use relevant features as privileged information. From 50 dimensions, a random index set $J$ of size 3 is sampled randomly. For all examples, feature values at these three indices are used as privileged information, i.e.,
$$x_i \sim \aleph(0, I_d)$$
$$x_i^* \leftarrow x_{iJ}$$
$$\alpha \sim \aleph(0, I_{d^*}), d^* = 3$$

For classification, the labels are generated by:
$$y_i \leftarrow \mathbb{I}(\alpha^T x_i > 0)$$

For regression, the targets are:
$$y_i \leftarrow \alpha^T x_i$$

*2.3.1.4. Sample-dependent Relevant Features as Privileged Information*

This experiment is similar to the previous one except that here 3 features are selected randomly from the input features for each sample to be used as privileged information. These three features determine the target value for each of the sample and are different for each example. That is, there are no globally important features, each the data is sampled as follows,
$$x_i \sim \aleph(0, I_d), d = 50$$
$$x_i^* \leftarrow x_{iJ_i}$$
$$\alpha \sim \aleph(0, I_{d^*}), d^* = 3$$

For classification,
$$y_i \leftarrow \mathbb{I}(\alpha^T x_i > 0)$$

For regression,
$$y_i \leftarrow \alpha^T x_i$$

### 2.3.2. MNIST Handwritten Digit Image Classification

We imitate the experiment carried out on MNIST handwritten digits [28] in [19] to further verify the effectiveness of our meta loss function for real-world classification using LUPI. LUPI is usually effective if privileged space is more informative than the regular input space and there is a relationship between the privileged and input spaces. Based on this premise, input and privileged sets are created from MNIST dataset in the same manner as done in [19]. The dataset comprises of 28x28 pixel images. The original images, being more informative, are used as privileged information and images down-scaled to 7x7 are used as input space examples. In this experiment, 500 examples are chosen randomly for training. All the three networks: teacher, input space and LUPI, consist of two hidden layers, the first with 16 rectified linear units, and the second with 32 rectified linear units. Testing is performed over the complete 10,000 image test dataset. Classification accuracy has been used as the performance evaluation metric in this case.

*2.3.3. Protein Binding Affinity Prediction*

To test the performance of our method in real-world regression problems, we applied it for prediction of protein binding affinity. Proteins are macromolecules responsible for carrying out most of the activities in cells of living organisms [29]. Proteins usually interact with each other and form protein complexes to perform these activities. These interactions depend on the binding affinity between them [30]. The study of protein interactions and binding affinity is an area of great interest for biologists and pharmaceutical experts due to its direct application in a number of biologically significant problems such as drug design, understanding cellular pathways, etc. [31]. However, experimental determination of binding affinity of proteins involved in an interaction for the formation for a protein complex is difficult [32]. Therefore, development of machine learning models for predicting binding affinity of proteins is a very active area of research [33]. Existing machine learning models for protein binding affinity prediction use either protein structure information [34] or protein sequences [35] for generating their predictions. 3D structures of proteins are more informative but are not available for every protein due to the significant cost and effort involved in protein structure determination [4]. On the other hand, sequence information, though not as informative, is easily available. As a consequence, development of a LUPI model that uses both sequence and structure information in training while requiring protein sequence information only in testing is ideally suited for this problem.

The task of binding affinity prediction is a regression problem with real-valued free energy values as targets, i.e, given a pair of proteins, predict the binding affinity between them. In an earlier study, we used large margin classification LUPI [6] for developing a protein binding affinity predictor for complexes [36]. We first remodeled the task as a classification problem. This was done by labeling complexes with target values less than -10.86 as positive class (+1). Although the nature of the problem had to be modified for using the classification LUPI solution, a considerable improvement in performance of the predictor as compared to the previous state-of-the-art was observed due to the use of structure-based features as privileged information and sequence features as the input space information.

In contrast to the previous LUPI classification approach, the proposed meta-loss function allows us to perform regression without changing the nature of the task to classification. The dataset used in this experiment is the protein binding affinity benchmark dataset version 2.0 [37]. It comprises of 144 non-redundant protein complexes with both sequence and 3D structure information available for each of the constituent proteins in a complex. For a given protein pair, 2-mer counts i.e., counts of all possible substrings of length 2 in both the proteins, have been used as input features. Moal Descriptors for representing protein structure as proposed by Moal et al. in [38] together with sequence features have been used as privileged information. Length of feature vectors for input and privileged space is 800 and 1000 respectively. Back-propagation Neural Neworks with one hidden layer have been used for both input and privileged space models. The hidden layers in both the nets comprise of the same number of neurons as the respective features space dimensions. Linear activation has been used in both the hidden and the output layers. LUPI model has the same architecture as the input space model. Leave One Complex Out cross-validation has been performed to evaluate our method. Root Mean Squared Error (RMSE), Spearman correlation coefficient, Area under the Precision-Recall and Receiver Operating Characteristic Curves (AUC-PR, AUC-ROC) have been used as performance metrics for evaluation. AUC-PR and AUC-ROC have been computed for comparing the method with the previous state-of-the-art classification based LUPI model [36] for the problem.

## 3. Results and Discussion

In this section, we present the performance evaluation results for the experiments described in the previous section.

*3.1. Synthetic Datasets*

Here we discuss the results for applying LUPI based on our formulation over artificially generated datasets. Four different cases, each representing a possible input-privileged space relation configuration have been analyzed. We have discussed each of the scenarios as follows.

*3.1.1. Clean Labels as Privileged Information*

As described earlier, in this experiment clean labels have been used as privileged space features and some noise is added to the given labels. We set $T = 0$ to ensure maximum knowledge transfer from the teacher. The results are presented in Table 1 and compared to the results reported in the work by Lopez-Paz et al over the same dataset [19]. It can be seen that LUPI based model outperforms the model trained over input space in both classification and regression settings. This shows that privileged information is helpful in improving the input space predictor for both classification and regression tasks as the privileged space information is more reliable in comparison to given labels. It is important to note that these results from the proposed formulation follow the same trend as the work by Lopez-Paz et al. [19]. In addition to a near perfect reconstruction of classification results in comparison to [19], the proposed scheme is able to perform LUPI based regression as well.

Table 1- Performance results on synthetic data using clean labels as privileged information

| **Feature Space** | **Lopez-Paz et al. [19]** | **Proposed meta-loss** | |
|---|---|---|---|
| | **Classification Accuracy** | **Classification Accuracy** | **Regression (RMSE)** |
| **Privileged** | $0.96 \pm 0.00$ | $0.95 \pm 0.01$ | $0.31 \pm 0.01$ |
| **Input** | $0.88 \pm 0.01$ | $0.87 \pm 0.01$ | $0.36 \pm 0.01$ |
| **LUPI** | $0.95 \pm 0.01$ | $0.95 \pm 0.01$ | $0.31 \pm 0.01$ |

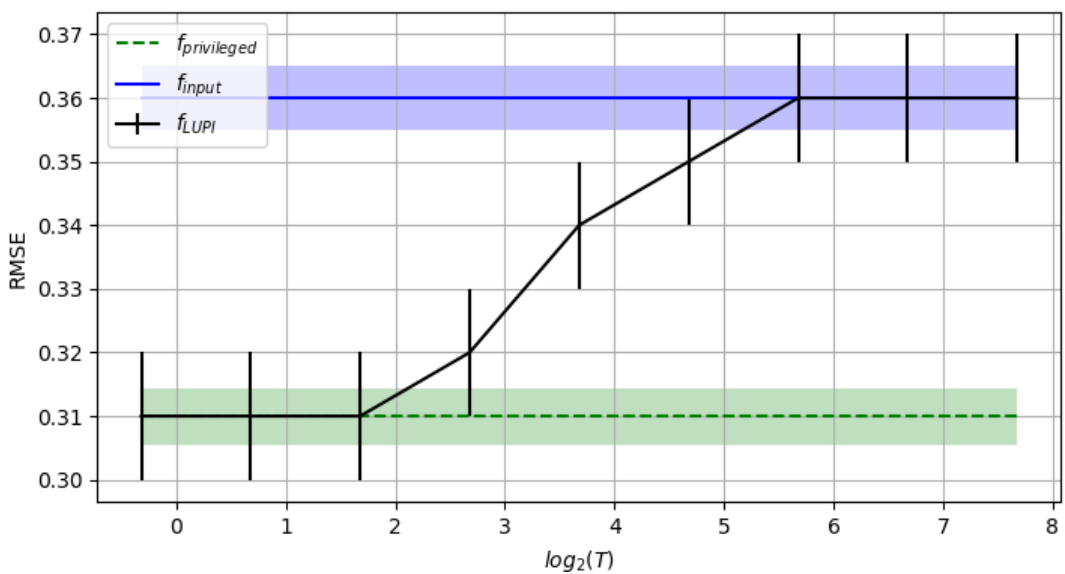

Figure 2- RMSE of the proposed model over varying values of T. For lower values of T, LUPI tends to mimic the teacher's behaviour. For high values, no knowledge is transferred from the teacher

To further analyse the behaviour of the proposed meta-loss over different values of the control parameter $T$ we present a plot of the mean and standard deviation of RMSEs for teacher, input-space and LUPI models in Figure 2. RMSE for privileged space is 0.31 and that for the input space is 0.36. It can be seen that, for small values of $T$, the LUPI based model copies the teacher model and for very high values, there is no knowledge transfer from the teacher, i.e., the performance of LUPI based model and simple input space model are almost same. This trend verifies the effective control of hyper-parameter $T$ on the extent of knowledge transfer from the teacher.

Table 2- Performance results when clean features are used as privileged information.

| Feature Space | Lopez-Paz et al. [19] | Proposed meta-loss | |
|---|---|---|---|
| | Classification Accuracy | Classification Accuracy | Regression RMSE |
| Privileged | 0.90 ± 0.01 | 0.89 ± 0.02 | 0.03 ± 0.01 |
| Input | 0.68 ± 0.01 | 0.68 ± 0.02 | 5.78 ± 0.59 |
| LUPI | 0.70 ± 0.01 | 0.68 ± 0.02 | 5.78 ± 0.59 |

*3.1.2. Clean Features as Privileged Information*

In this experiment, noise has been added to the input samples and clean features have been used as privileged information. The results for this experiment are presented in Table 2. Since the additive feature noise is independent of the privileged information, no useful knowledge can be transferred. Therefore, no improvement in the input space performance has been observed. However, it is interesting to note that the LUPI based model gives the same level of classification and regression performance as the input space model. Here again, the trend for both classification and regression based on our formulation is the same as that presented in [19].

*3.1.3. Relevant Features as Privileged Information*

For this experiment, only relevant features are used as privileged information. We present the results for this experiment in Table 3. A considerable improvement in classification performance is seen for learning using privileged information. However, in case of regression, the problem becomes very easy to learn even for the input space classifier since the privileged information is a subset of the input features and labels are directly dependent on the privileged information. So, RMSE for the input space is also very low and not much improvement can be seen due to application of LUPI. The problem is relatively harder in case of classification due to binarization of labels. Therefore, the usefulness of LUPI is more prominent in classification settings for this problem. We have also tested the performance of predicting classification labels using Mean Squared Error (MSE) as a loss function which further illustrates this point (see column 4 in Table 3).

Table 3- Performance results when relevant features are used as privileged information.

| Feature Space | Lopez-Paz et al. [19] | Proposed meta-loss | | |
|---|---|---|---|---|
| | Classification Accuracy | Classification Accuracy | Regression RMSE | Regression over binary labels RMSE |
| Privileged | 0.98 ± 0.00 | 0.99 ± 0.01 | 0.01 ± 0.00 | 0.30 ± 0.00 |
| Input | 0.89 ± 0.01 | 0.87 ± 0.03 | 0.14 ± 0.04 | 0.38 ± 0.02 |
| LUPI | 0.97 ± 0.01 | 0.98 ± 0.01 | 0.14 ± 0.03 | 0.34 ± 0.02 |

*3.1.4. Sample-dependent Relevant Features as Privileged Information*

In this experiment, features relevant for each of the example were passed as privileged information. Although the problem can be modelled linearly in privileged space, there is no globally important information that can be transferred from the teacher to student. Therefore, LUPI in this case does not help in improving performance. Results are presented in Table 4.

Table 4- Performance results for Sample-dependent relevant features used as privileged information

| Feature Space | Lopez-Paz et al. [19] | Proposed meta-loss | |
|---|---|---|---|
| | Classification Accuracy | Classification Accuracy | Regression RMSE |
| Privileged | $0.96 \pm 0.02$ | $0.97 \pm 0.02$ | $0.01 \pm 0.00$ |
| Input | $0.55 \pm 0.03$ | $0.50 \pm 0.03$ | $2.12 \pm 1.01$ |
| LUPI | $0.56 \pm 0.04$ | $0.51 \pm 0.01$ | $2.12 \pm 1.01$ |

Performance results in the above-mentioned experiments show that distillation using privileged information based teacher model is effective whenever a relationship between the two spaces exists and the privileged information is more informative in comparison to input space features. Furthermore, these classification results are in agreement with the work by Lopez-Paz et al. [19]. However, the proposed scheme can also perform regression and can possibly be used for other machine learning problems as well with an appropriate loss function.

*3.2. MNIST Handwritten Digit Image Classification*

This experiment was performed using 500 randomly chosen training examples from MNIST dataset [28]. Original, full resolution images were used as privileged space examples and their down-scaled versions as input space examples. Testing is performed over complete 10,000 image dataset. The results are presented in Table 5. Classification accuracy using both our proposed formulation and formulation in [19] has been presented. Our method produces a similar improvement in LUPI as in [19]. It can be seen that, given the privileged net is trained properly, i.e., it does not over/under-fit, LUPI improves performance. In this case, the use of privileged information in training has produced an improvement of 9% classification accuracy. These results clearly show that the proposed scheme is as effective as the original classification formulation by Lopez-Paz et al.

Table 5- Classification Accuracy for MNIST dataset

| Feature Space | Lopez-Paz et al. [19] | Proposed Formulation |
|---|---|---|
| Privileged | 0.84 | 0.81 |
| Input | 0.59 | 0.67 |
| LUPI | 0.75 | 0.76 |

*3.3. Protein Binding Affinity Prediction*

In this experiment, we have used the proposed LUPI formulation for a practical regression problem: prediction of protein binding affinity. For comparison, we have used our previous work which uses protein sequence features as input space features and protein structure information as privileged features for a classification based modeling of the prediction problem to give state of the art classification results [36]. In this work, we have used both classification and regression based LUPI formulations. The results are presented in Table 6. It is interesting to note that LUPI gives a significant improvement in both classification and regression tasks in comparison to input features alone. Furthermore, it can be seen that the proposed regression based LUPI outperforms classification LUPI in all performance metrics. An improvement of 7% in AUC-PR and 1% in AUC-ROC has been observed as compared to the LUPI-SVM solution of the problem. A major improvement in Spearman Correlation from 0.48 to 0.56 has also been observed. This clearly shows that the proposed regression based modeling of the problem is very effective in real-world applications.

Table 6- Leave One Complex Out cross-validation results for Binding Affinity Problem. Bold values indicate optimal performance.

| Method | Space | AUC-ROC | AUC-PR | Correlation | RMSE |
|---|---|---|---|---|---|
| LUPI for classification [36] | Input | 0.72 | 0.68 | 0.40 | - |
| | Privileged | 0.73 | 0.68 | 0.43 | - |
| | LUPI | 0.78 | 0.73 | 0.48 | - |
| Proposed Regression LUPI | Input | 0.70 | 0.70 | 0.41 | 3.14 |
| | Privileged | 0.75 | 0.75 | 0.50 | 3.07 |
| | LUPI | **0.79** | **0.80** | **0.56** | **3.03** |

## 4. Conclusions

In this work, we have addressed a limitation of existing large margin and neural network formulations of learning using privileged information (LUPI) that are restricted to classification only by proposing a simple yet effective change in the meta-loss function used in the structural risk minimization formulation of LUPI that allows it to model both regression and classification problems. The proposed formulation is generalized in that it can be easily extended to model other machine learning problems such as ranking and recommender systems. In contrast to previous formulations, the proposed meta-loss is relatively simpler with a single control hyper-parameter to regulate the contribution of privileged space and performing softening. We have demonstrated correctness and effectiveness of the proposed loss by performing several classification and regression experiments on both synthetic and real-world datasets. Our LUPI model has shown to outperform state of the art predictor of protein binding affinity in complexes. Our work opens the avenues for further application of learning using privileged information and distillation in various machine learning problems. The proposed scheme has also the potential to be utilized for distillation problems.

**Acknowledgements**

We are grateful to Dr. David Lopez-Paz, Facebook AI Research, Paris, France for his valuable input and suggestions regarding the mathematical formulation and experimental analysis.

AA and MD are funded via Information Technology and Telecommunication Endowment Fund at Pakistan Institute of Engineering and Applied Sciences.

**Conflict of Interest**

The authors declare no conflict of interest.